%% file: main.tex
\documentclass[runningheads,a4paper]{llncs}

\usepackage{times}
\usepackage{epsfig}
\usepackage{graphicx}
\usepackage{amsmath}
\usepackage{amssymb}
\usepackage{xspace}
\usepackage[numbers,sort]{natbib}
\usepackage{booktabs}
\usepackage{multirow}
\usepackage{pgfplots}
\pgfplotsset{compat=newest}

\let\oldbibliography\bibliography% Store \bibliography in \oldbibliography
\renewcommand{\bibliography}[1]{{%
  \let\chapter\section% Copy \section over \chapter
  \oldbibliography{#1}}}% Old \bibliography

\newcommand{\set}[1]{\ensuremath{\mathcal{#1}}}
\newcommand{\con}[1]{#1} %\ensuremath{\mathsf{#1}}}

\newcommand{\argmax}{\operatornamewithlimits{\arg\,\max}}

\newcommand{\diag}[1]{\mathrm{diag}(#1)}

\def\eg{\textit{e.g.}~}
\def\ie{\textit{i.e.}~}
\def\Eg{\textit{E.g.}~}
\def\etal{\textit{et al.}~}

\newcommand{\DAL}{DA-layer\xspace }
\newcommand{\DALs}{DA-layers\xspace }
\newcommand{\DIAL}{DIAL\xspace }
\newcommand{\DIALAlexNet}{DIAL -- AlexNet\xspace }
\newcommand{\DIALInception}{DIAL -- Inception-BN\xspace }

\begin{document}

\mainmatter

\title{Just DIAL: DomaIn Alignment Layers for Unsupervised Domain Adaptation}
\titlerunning{Just DIAL}  % abbreviated title (for running head)
\author{%
Fabio Maria Carlucci\inst{1}\and%
Lorenzo Porzi\inst{2}\and%
Barbara Caputo\inst{1}\\%
Elisa Ricci\inst{3,4}\and%
Samuel Rota Bul\`o\inst{3}
}
\authorrunning{Fabio Maria Carlucci et al.} % 

\institute{
Sapienza University, Roma, Italy \and
Institut de Robotica i Informatica Industrial CSIC-UPC, Barcelona, Spain \and
Fondazione Bruno Kessler, Trento, Italy \and 
University of Perugia, Perugia, Italy
}

\date{}

\tocauthor{Fabio Maria Carlucci, Lorenzo Porzi, Barbara Caputo, Elisa Ricci and Samuel Rota Bul\`o}

\maketitle              % typeset the title of the contribution

\begin{abstract}
The empirical fact that classifiers, trained on given data collections, perform poorly when tested on data acquired in different settings is theoretically explained in domain adaptation through a shift among distributions of the source and target domains.
Alleviating the domain shift problem, especially in the challenging setting where no labeled data are available for the target domain, is paramount for having visual recognition systems working in the wild.
As the problem stems from a shift among distributions, intuitively one should try to align them.
In the literature, this has resulted in a stream of works attempting to align the feature representations learned from the source and target domains
by introducing appropriate regularization terms in the objective function.
In this work we propose a different strategy and we act directly at the distribution level by introducing \emph{DomaIn Alignment Layers} (\DIAL) which reduce
the domain shift by matching the source and target feature distributions to a canonical one.
Our experimental evaluation, conducted on a widely used public benchmark, demonstrates the advantages of the proposed domain adaptation strategy. %method on three different public benchmarks we confirm the power of our approach.

\end{abstract}
\vspace{-0.5cm}
\section{Introduction}
\vspace{-0.2cm}
\label{intro.tex}
\input{intro}
\vspace{-0.2cm}
\section{Related Work}
\vspace{-0.2cm}
\label{related}
\input{related.tex}

%\section{The Method}
\vspace{-0.3cm}
\section{DIAL: DomaIn Alignment Layers}
\vspace{-0.2cm}
\label{method}
\input{method}

\input{experiments}
\input{results}
\vspace{-0.2cm}
\section{Conclusions}
\vspace{-0.2cm}
\label{conclu}
In this paper we presented DIAL, a general framework for unsupervised, deep domain adaptation.
Our main contribution is the introduction of novel, domain-alignment layers, which reduce domain shift by matching source and target distributions to a freely definable reference distribution. We also show that improved performance can be
obtained by exploiting unlabeled target data introducing an entropy loss in the objective function.
We evaluated the proposed approach devising a simple implementation of our \DALs based on multiple batch normalization transformations.
The results of our experiments demonstrate that DIAL outperforms state-of-the-art domain adaptation methods.
%In this work we focused on the challenging problem of unsupervised domain-adaptation and considered a single source/single task scenario, but our approach can be trivially extended to a semi-supervised setting and to multiple domains. 
Future works will investigate how to extend the proposed approach to a multi-source/multi-target setting. We also plan to consider other reference distributions for domain alignment in order to further improve performance. %  suit the domain adaptation task. 
\vspace{-0.4cm}
\bibliographystyle{splncs}
\bibliography{egbib}

\end{document}

%% file: intro.tex
\begin{figure}
  \centering
  \includegraphics[width=0.6\columnwidth]{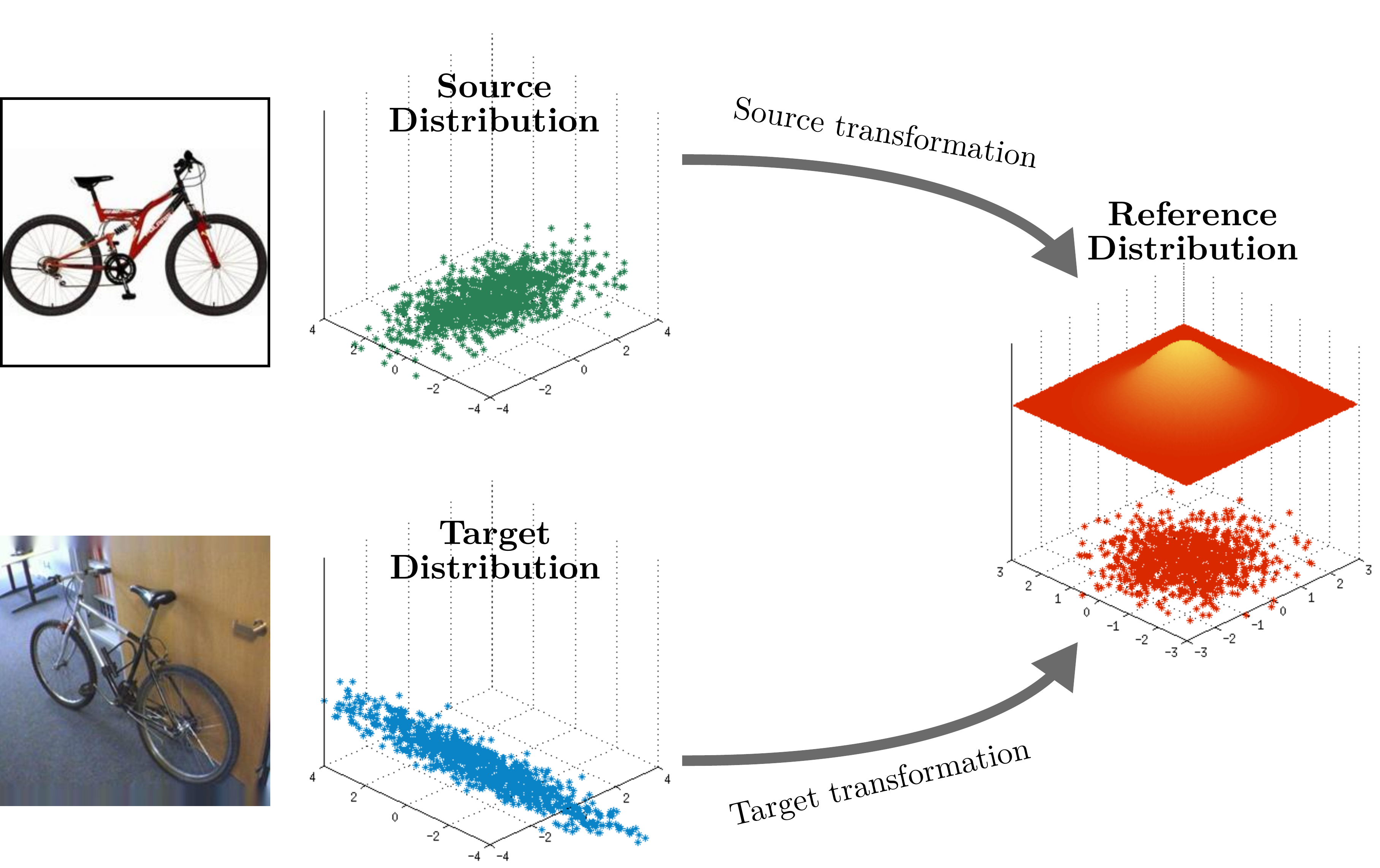}
  \caption{\DIAL learn a pair of transformations that shift the observed source and target distribution to match a desired reference distribution.}
  \label{fig:teaser}
  \vspace{-0.7cm}
\end{figure}

Many scientists today believe we are witnessing the golden age of computer vision.
The massive adoption of machine learning and, in particular, of deep learning techniques as well as the availability of large fully annotated datasets have enabled 
amazing progresses in the field.
A natural question is if the novel generation of computer vision technologies is robust enough to operate in real world scenarios.
One of the fundamental requirements for developing systems working in the wild is devising computational models which are immune to the domain shift problem, 
\ie which are accurate when test data are %to the challenge that usually training and test data are not independently and identically 
drawn from a (slightly) different data distribution than training samples. 
Unfortunately, recent studies in the literature have shown that, even with powerful deep architectures, the domain shift problem
can only be alleviated but not entirely solved \cite{donahue2014decaf} and several methods for deep domain adaptation have been developed.

Domain adaptation focuses on learning classification or regression models on some target data by exploiting additional knowledge derived
from a related source task.
In particular, unsupervised domain adaptation focuses on the challenging scenario where no labeled data are available in the target domain. % 
%deserves special attention, as in many applications annotating data is not only a tedious operation but may be not possible.
Several approaches have been proposed for unsupervised domain adaptation in the past, the most successful of which are %both considering hand-crafted 
%features~\cite{huang2006correcting,gong2013connecting,gong2012geodesic,long2013transfer,fernando2013unsupervised} and 
based on deep architectures~\cite{long2015learning,tzeng2015simultaneous,ganin2014unsupervised,long2016unsupervised}.
%Among recent works based on convolutional neural networks (CNN), feature alignment methods have achieved remarkable performance. Feature alignment approaches aim to discover a new feature representation for the source and target domain in order to reduce the discrepancy among data distributions. Two strategies are typically employed for aligning source and target distributions. One strategy is based on the minimization of Maximum Mean Discrepancy (MMD) \cite{long2015learning,long2016unsupervised}, \ie additional layers are introduced in the network to force the distributions of the learned source and target representations to be as similar as possible in MMD terms. A second strategy \cite{tzeng2015simultaneous,ganin2014unsupervised} is based on the introduction of a domain-confusion loss. This loss is used to learn an auxiliary classifier predicting if a sample comes from the source or the target domain. Intuitively, by maximizing this term, \ie by imposing the auxiliary classifier to exhibit poor performance, domain-invariant features are learned.
%In particular, recent works based on convolutional neural networks (CNN) have achieved remarkable performance.
%Most of these methods attempt to reduce the discrepancy among source and target distributions by learning features that are invariant to the domain shift.
Previous unsupervised domain adaptation methods can be roughly divided in two categories. The first category includes methods which attempt to reduce
the discrepancy between source and target distributions by minimizing the distance between the mean embeddings of the learned representations, \ie the so-called
Maximum Mean Discrepancy (MMD)~\cite{long2015learning,long2016unsupervised}. A second class of methods learns domain invariant features by maximizing
a domain-confusion objective function, modelling the loss of an auxiliary classifier which should discriminate if a sample belongs to 
the source or to the target domain \cite{tzeng2015simultaneous,ganin2014unsupervised}.

Following these recent approaches, in this paper we present a domain adaptation method which simultaneously learns discriminative deep representations 
while coping with domain shift in the unsupervised setting.
Differently from previous works, we do not focus on learning domain-invariant features by \emph{explicitly} optimizing additional loss terms (\eg MMD, domain-confusion).
We argue instead %that different predictors should be used for source and target samples and 
that domain adaptation can be achieved
by embedding in the network some \textit{Domain Alignment} layers (\DALs) which operate by aligning both source and target distributions to a canonical one. We also show that several different transformations can be employed in our \DALs to match source and target data distributions to the reference, thus highlighting the generality of our approach.
%\elisa{aligning the source and target distributions to predefined reference distribution} (Fig.\ref{fig:teaser}).
%\elisa{Discuss all transformations}
%We also show that with our framework
%to exploit unlabeled target data by introducing an entropy loss in the objective function, such as to promote
%to construct a prior distribution on the network parameters, biasing the learned solution towards 
%models that maximally separate the target classes. % in the target domain.
%Finally, we show how, under certain conditions, our approach can be completely implemented by employing traditional layers, \ie batch normalization (BN, \cite{ioffe2015batch}) and the cross-entropy loss.
We call our algorithm DIAL -- DomaIn Alignment Layers.
%Our extensive experimental evaluation, conducted on publicly available datasets,
Our experimental evaluation, conducted on the most widely used domain adaptation benchmark, \ie the \textbf{Office-31} \cite{saenko2010adapting} dataset,
demonstrates that DIAL greatly alleviates the domain discrepancy 
and outperforms most state of the art techniques.

%% file: related.tex
In the last decade unsupervised domain adaptation %focuses on the 
%challenging 
%scenario where
%labeled data are only available in the source domain. 
have received considerable interest in the computer vision community as in many applications labeled data are not available in the target domain \cite{huang2006correcting,chu2013selective,yamada2012no,gong2013connecting,zeng2014deep,zen2014unsupervised,costante2013transfer,long2015learning,tzeng2015simultaneous,ganin2014unsupervised}.

Traditional methods for unsupervised domain adaptation attempt to reduce the domain shift
by adopting two main approaches.
%addressed the problem of reducing the discrepancy between the source
%and the target distributions by considering two main strategies. 
A first strategy, the so-called instance re-weighting \cite{huang2006correcting,chu2013selective,yamada2012no,gong2013connecting,zeng2014deep}, is based
on building models for the target domain by adopting appropriately re-weighted source samples. The idea is to assign different importance to 
source samples such as to reflect their similarity with the target data. 
This approach has been proposed in \cite{huang2006correcting} where a nonparametric
method called Kernel Mean Matching is used to set weights without explicitly estimating the data distributions.
Similarly, Gong \etal\cite{gong2013connecting} introduced the notion of landmark datapoints, a
subset of source samples which are similar to target data, and proposed a landmark-based domain adaptation method.
Chu \etal \cite{chu2013selective} presented a framework for joint source sample selection
and classifier learning. % adopting within a single optimization problem. 
While these works considered hand-crafted features, similar ideas can be also exploited in the 
case of deep architectures. An example is the work in \cite{zeng2014deep} where deep 
autoencoders are used to build source sample weights.
%an unsupervised domain adaptation approach for pedestrian detection 
%and used 
%using deep autoencoders to weight the importance of source training samples.
%and applied to the problem of learning personalized model in the context of facial expression analysis.

The large majority of previous unsupervised domain adaptation methods are based on
feature alignment, \ie domain shift is reduced by projecting source and target data 
in a common subspace. Several feature alignment methods have been 
proposed in the past, both considering shallow models \cite{gong2012geodesic,long2013transfer,fernando2013unsupervised} and deep architectures 
\cite{long2015learning,tzeng2015simultaneous,ganin2014unsupervised}.
Focusing on works adopting deep architectures, most methods align source and target feature representations by adding
in the objective function a regularization term attempting to (i)
reduce Maximum Mean Discrepancy \cite{long2015learning,long2016unsupervised,sun2016deep} or (ii) maximize a domain confusion 
loss \cite{tzeng2015simultaneous,ganin2014unsupervised}.
Recent studies have also investigated alternative methodologies, such as building specific
encoder-decoder networks to jointly learn source labels and reconstruct unsupervised target images
~\cite{ghifary2016deep,bousmalis2016domain}.
Our approach significantly departs from previous works by reducing the discrepancy between 
source and target distributions through the introduction of our DA-layers.
The most similar work to ours is \cite{li2016revisiting} where Li \etal proposed to revisit batch normalization for deep
domain adaptation. While our approach develops from a similar  
intuition, our method
can be regarded as a generalization of \cite{li2016revisiting}, as we consider several transformation in our DA layers and 
we 
%also 
introduce a prior over the network parameters in order to benefit from the target samples during training.
Experiments presented in Sec.~\ref{experiments} show the significant added value of our idea.
% CORAL It minimizes the difference between distributions of source and
% target features by aligning the second-order statistics, namely,
% the covariance. But a hyper-parameter is required to guarantee
% the stability of the algorithm.

%% file: method.tex
Let $\set X$ and $\set Y$ denote the input space (\eg images) and the output space (\eg image categories) of our learning task, respectively.
We consider an unsupervised domain adaptation setting, where we have a \emph{source domain} described in terms of a probability distribution $p^s_{\mathtt xy}$ over $\set X\times\set Y$ and a \emph{target domain} following $p^t_{\mathtt xy}$.
The source and target distributions differ in general and are unknown, but we are provided with $\con n$ labeled observations $\set S=\{(x_1^s,y_1^s),\ldots,(x_{\con n}^s,y_{\con n}^s)\}$ from the source domain, \ie they are sampled from $p^s_{\mathtt xy}$, and $\con m$ unlabeled observations $\set T=\{x_1^t,\ldots,x_{\con m}^t\}$ sampled from the marginal distribution $p^t_{\mathtt x}$. The goal of the learning task is to estimate a predictor for the target domain, using the observations in $\set S$ and $\set T$. This task is particularly challenging because we lack observed labels from the target domain and the discrepancy between the source and target domains, which in general exists, prevents predictors trained on the source domain to be readily applicable to samples from the target domain.

One key element for the success of an unsupervised domain adaptation algorithm is its ability of reducing the discrepancy between source and target domains. There are different approaches to achieve this goal, but we focus on aligning the domains at the feature level. Within this family of methods the most successful ones \emph{couple} the training process and the domain adaptation step within \emph{deep} neural architectures~\cite{ganin2014unsupervised,long2016unsupervised,long2015learning}, yielding alignments at different level of abstractions. Our method is close in spirit to this line of works but we distinguish from them by \emph{(a)} not relying on the covariate shift assumption, \ie we in general assume $p^s_{\mathtt y|x}\neq p^t_{\mathtt y|x}$, and by \emph{(b)} hard-coding the domain-invariance properties directly into our deep neural network. The rationale behind the former choice is the impossibility theorem for domain adaptation given in~\cite{ben2010impossibility}, which intuitively states that no domain adaptation algorithm can succeed (in terms of the notion of learnability) if it relies on the covariate shift assumption and achieves a low discrepancy between the source and target unlabeled distributions, \ie $p^s_{\mathtt x}$ and $p^t_{\mathtt x}$, respectively. Since the latter assumption is what one implicitly pursues by performing domain alignments at the feature level, we drop the former assumption. The other distinguishing aspect of our method is an architectural solution to achieve domain-invariance, which contrasts with the majority of approaches that rely on additional loss terms (\eg MMD-type losses~\cite{long2015learning} or adversarial losses \cite{tzeng2015simultaneous,ganin2014unsupervised}) that induce an \emph{external} pressure on the networks' parameters at training time to fulfill the domain-invariance requirement. 
Works exists that do not rely on the covariate-shift assumption and take a loss-based approach to feature alignment, but those typically implement the source and target predictors using different sets of parameters (not necessarily disjoint)~\cite{rozantsev2016beyond,long2016unsupervised}. Instead, the method we propose is able to avoid the covariate shift assumption and at the same time have the set of learnable network parameters, denoted by $\theta$ in this work, that is totally shared. The key element of our method is the 
%so-called 
domain-alignment layer that we describe below.
\vspace{-0.4cm}
\subsection{Source and Target Predictors}
\label{sec:predictors}
\vspace{-0.2cm}
We implement source and target predictors as two deep neural networks that share the same structure and the same parameters given by $\theta$. However, the two networks differ by having a number of layers that perform a domain-specific operation. Those layers are called \emph{Domain-Alignment Layers} (\DALs) and their role is to apply a data transformation that aligns the input distribution to a pre-fixed reference distribution. In Figure~\ref{fig:teaser}, we provide an illustration of the basic principle. In general, the input distributions to \DALs
in the source and target predictors differ, but the reference distribution remains fixed. As a result, the data transformations that are applied in the \DALs of the source and target predictors differ. Consequently, source and target predictors implement different functions, thus violating the aforementioned covariate shift assumption, while still sharing the same set of learnable parameters.
More details about the neural network architectures will be provided in the experimental section.

To better understand how the domain-alignment transformation works, we consider a single \DAL in isolation.
The desired output distribution, namely the reference distribution, is decided a priori and thus known.
The input distribution instead is unknown, but we can rely on a sample $\set D$ thereof. Now given a transformation $g$ from a family of transformations $\set G$ we can push the reference distribution into the pre-image under $g$ via a variable change. This yields a family of distributions
among which we can select the one, say $\hat g$, that most likely represents sample $\set D$.
%yielding a distribution parametrized by $g$ that could be fit to sample $\set D$ via maximum likelihood estimation. 
In other words, if $\mathtt v$ is a random variable following the reference distribution and we assume that the input observations in $\set D$ are realizations of random variable $\mathtt u=g^{-1}(\mathtt v)$, then we can determine the transformation $\hat g\in\set G$ as the one that maximizes the likelihood $p_\mathtt u(\set D|g)$. 
We can alternatively encode some prior knowledge about the transformation by taking a Maximum-A-Posteriori (MAP) approach and thus maximize $p_\mathtt u(g|\set D,\psi)$, where $\psi$ encodes hyper-parameters governing the prior over $g$.

This idea paves the way to a number of transformations that could be obtained by playing with different reference distributions and families of transformations. In this work, we restrict our focus to some families of \DALs.
In all the cases we consider in this work we assume that $\set G$ consists of channel-wise linear transformations of the form $\set G=\{u\mapsto \diag a^{-\frac{1}{2}}(u-b)\,:\,a, b\in\mathbb R^d, a>0\}$ where $\diag a$ is a diagonal matrix with diagonal elements given by $a$.
A first family of approaches is obtained by imposing the standard normal distribution as reference distribution and depending on the prior knowledge we inject we obtain the following variations of \DALs:
\vspace{-0.2cm}
\paragraph{Batch normalization.}
By pushing the standard normal distribution, \ie the reference distribution of $\mathtt v$, into the pre-image under $g\in\set G$ we obtain a distribution for random variable $\mathtt u=g^{-1}(\mathtt v)$ that is normal with mean $b$ and covariance $\diag a$. The maximum likelihood estimates of $a$ and $b$ given sample $\set D$, consisting of i.i.d. realizations of $\mathtt u$, are given by $\hat a=\sigma^2 (\set D)$ and $\hat b=\mu(\set D)$ respectively, where $\mu(\set D)$ and $\sigma^2(\set D)$ represent the sample mean and the diagonal of the sample covariance of $\set D$, respectively. The resulting domain-alignment transformation is $\hat g(u)=\diag{\sigma^2(\set D)}^{-\frac{1}{2}}\left[ u-\mu(\set D) \right]$. This transformation corresponds to the well-known \emph{batch-normalization} layer~\cite{ioffe2015batch}, when $\set D$ is the mini-batch of a training iteration. 
\vspace{-0.2cm}
\paragraph{Batch normalization with prior on variance.}
This setting is similar to the previous one, but instead of considering a maximum likelihood estimate of the transformation parameter $a$ we opt for a MAP estimate. To this end we introduce an Inverse-Gamma($\alpha$,$\beta$) prior on the transformation parameter $a$, yielding a posterior distribution for $a$ that is Inverse-Gamma($\bar\alpha$,$\bar\beta$) with $\bar\alpha=\alpha+\frac{|\set D|}{2}$ and $\bar\beta=\beta +\frac{|\set D|}{2}\sigma^2(\set D)$. The corresponding MAP estimate is given by $\hat a=\frac{\bar\beta}{\bar\alpha+1}$. The hyperparameters of the prior distribution, namely $\alpha$ and $\beta$ are set to $\alpha=\frac{|\set D|}{2}-1$ and $\beta=\epsilon\frac{\set D}{2}$, where $\epsilon$ is intuitively a prior variance. In this way we have that $\hat \beta$ gives approximately equal weight to the sample variance and the prior variance, yielding $\hat\beta =\epsilon + \sigma^2(\set D)$. Finally, the estimate of $b$ remains the maximum likelihood estimate, namely the sample mean, \ie $\hat b=\mu(\set D)$. Note that the data transformation we obtain with this procedure is the actual implementation of batch normalization that we find in most deep learning frameworks, for $\epsilon$ typically appears as a small additive constant for the variance that prevents numerical issues. In our case, however $\epsilon$ is not necessarily set to a small constant as we will see in the experimental section.

%In this setting we assume that the input sample to the layer is affected by independent Gaussian noise with zero mean and covariance $\diag{\epsilon\vct 1}$, where $\vct 1$ is a vector of $1$s and $\epsilon > 0$.
%In other terms, we assume $\mathtt u=g^{-1}(\mathtt v)+\mathtt n$ 
%where $\mathtt v$ follows the standard normal distribution (reference distribution) and $\mathtt n$ is a $d$-dimensional vector of normal variates $\mathcal N(\mu=0,\sigma^2=\epsilon)$. Accordingly, $\mathtt u$ is a Gaussian random vector with mean $b$ and covariance $\diag{a+\epsilon\vct 1}$. Under this assumption the maximum likelihood estimates of $a$ and $b$ are $a=\mu(\mathcal D)$ and $b=\sigma^2(\mathcal D)+\epsilon\vct 1$, respectively. Note that this is the actual implementation of batch normalization that we find in most deep learning frameworks, for $\epsilon$ typically appears as a small additive constant for the variance that prevents numerical issues. 

A second family of approaches is obtained by imposing the Laplace distribution as reference distribution. In this case we do not explore variations involving prior knowledge, although it would be possible.
\vspace{-0.2cm}
\paragraph{Laplace Batch normalization.}
If we assume that the reference distribution follows a standard Laplace distribution, then the maximum likelihood estimate $\hat b$ corresponds to the sample median, while the maximum likelihood estimate of $a$ is given by the mean absolute value deviation from the sample median, \ie $\hat a=\frac{1}{|\set D|}\sum_{x\in\set D}|x-\hat b|$.

\vspace{-0.5cm}
\subsection{Training and Inference}
\vspace{-0.2cm}
\paragraph{Training.} During the training phase we consider the datasets $\set S$ and $\set T$ and we estimate the neural network weights $\theta$. Note that these parameters are shared by the source and the target predictors.
To compute $\theta$ we define a posterior distribution of $\theta$ given the observations $\set S$ and $\set T$, $\pi(\theta|\set S,\set T)$, and maximize it over $\Theta$ to obtain a MAP estimate $\hat\theta$:
\begin{equation}
    \hat\theta \in\argmax_{\theta\in\Theta} \pi(\theta|\set S,\set T)\,.
    \label{eq:MAP}
\end{equation}
The posterior distribution is defined as $\pi(\theta|\set S,\set T)\propto\pi(y_{\set S}|x_{\set S},\theta)\pi(\theta|\set T)$, where $y_{\set S}=\{y_1,\dots,y_n\}$ and $x_{\set S}=\{x_1,\dots,x_n\}$ indicate the set of labels and data points in $\set S$, respectively. 
The term $\pi(y_{\set S}|x_{\set S},\theta)$ is the likelihood of $\theta$ with respect to the source dataset,
while
%given a sample $(x,y)$ from the source domain. This corresponds to \eg the $y$th output element of a neural network having a soft-max layer on top. The prior term 
$\pi(\theta|\set T)$ is a prior term depending on the unlabeled target samples.
%which acts as a regularizer in the classical learning theory sense.
Assuming the data samples to be \emph{i.i.d.}, the likelihood term is given by
\begin{equation}
    \pi(y_{\set S}|x_{\set S},\theta)=\prod_{i=1}^nf^\theta_{y^s_i}(x^s_i;x_{\set S})\,,
    \label{eq:likelihood}
\end{equation}
where $f^\theta_{y^s_i}(x^s_i;x_{\set S})$ is the probability that sample point $x^s_i$ takes label $y^s_i$ according to the source predictor.
%The likelihood $\pi(y_i|x_i,x_{\set S},\theta)$ related to each sample point $(x_i,y_i)\in\set S$ still depends on the source dataset for reasons we clarify later.

%Before delving into the details of the prior term, we would like to remark on the absence of an explicit component in the probabilistic model that tries to align the source and target distributions.
%This is because in our model the domain-alignment step is taken over by each predictor, independently, via the domain-alignment layers as shown in the previous subsection.
In analogy to previous works on semi-supervised learning ~\cite{grandvalet2004semisupervised} and unsupervised domain adaptation \cite{long2016unsupervised}, the prior distribution $\pi(\theta|\set T)$ is defined in order to promote models that exhibit well separated classes.
This is achieved by defining $\pi(\theta|\set T)\propto\exp\left( -\lambda\, h(\theta|\set T) \right)$, where $\lambda$ is a user-defined parameter and $h(\theta|\set T)$ is the empirical entropy of $\mathtt y|\theta$ conditioned on $\mathtt x$, \ie:
\begin{equation}
    h(\theta|\set T)=-\frac{1}{m}\sum_{i=1}^m\sum_{y\in\set Y}f_y^\theta(x^t_i;x_\set T)\log f_y^\theta(x^t_i;x_\set T)\,,
    \label{eq:h}
\end{equation}
where $f_y(x^t_i;\set T)$ represents the probability that sample point $x_i^t$ takes label $y$ according to the target predictor. %In this way the predictor is based on the degree of label uncertainty that is observed when the same predictor is applied to the target samples.

%This prior distribution has been introduced in the context of semi-supervised learning in~\cite{grandvalet2004semisupervised} and % and employed for unsupervised domain adaptation in~\cite{long2016unsupervised}.
%The resulting prior distribution   satisfies the desired property of preferring models that exhibit well separated classes (\ie having lower values of $h(\theta|\set T)$), thus enabling the exploitation of the information content of unlabeled target observations within a discriminative setting~\cite{grandvalet2004semisupervised}.

\paragraph{Inference.}
Once the optimal network parameters $\hat\theta$ are estimated by solving~\eqref{eq:MAP}, the dependence of the target predictor $f_y^\theta(x;x_\set{T})$ from $x_\set{T}$ can be removed.
In fact, after fixing $\hat\theta$, the input distribution to each \DAL also becomes fixed, and we can thus compute and store the required transformation once and for all. \Eg, for the special case of \emph{Batch normalization} discussed in Section~\ref{sec:predictors}, this means simply to store the values of $\mu(\set D)$ and $\sigma(\set D)$.

%% file: experiments.tex
\vspace{-0.2cm}
\section{Experiments}
\vspace{-0.2cm}
\label{experiments}
In this section we extensively evaluate our approach and compare it with state-of-the-art unsupervised domain adaptation methods. We also provide a detailed analysis of the proposed framework, performing a sensitivity study and demonstrating empirically the effect of our domain alignment strategy.
\vspace{-0.4cm}
\subsection{Experimental Setup}
\vspace{-0.2cm}
% \subsubsection{Datasets} We evaluate the proposed approach on three publicly-available datasets.

To evaluate the proposed approach, we consider the \textbf{Office-31} \cite{saenko2010adapting} dataset.
Office-31 is a standard benchmark for testing domain adaptation methods.
It contains 4652 images organized in 31 classes from three different domains: Amazon (A), DSRL (D) and Webcam (W).
Amazon images are collected from \texttt{amazon.com}, Webcam and DSLR images were manually gathered in an office environment.
In our experiments we consider all possible source\slash{}target combinations of these domains and adopt the \emph{full protocol} setting \cite{gong2013connecting}, \ie we train on the entire labeled source and unlabeled target data and test on annotated target samples.

%The \textbf{Office-Caltech}\cite{gong2012geodesic} dataset is obtained by selecting the subset of $10$ common categories in the Office31 and the Caltech256\cite{griffin2007caltech} datasets.
%It contains $2533$ images of which about half belong to Caltech256.
%Each of Amazon (A), DSLR (D), Webcam (W) and Caltech256 (C) are regarded as separate domains.
%In our experiments we only consider the source\slash{}target combinations containing C as either the source or target domain.
\vspace{-0.5cm}
\subsubsection{Networks and Training.}
\label{sec:setup}

We apply the proposed method to two state-of-the-art CNNs, \ie AlexNet~\cite{krizhevsky2012imagenet} and Inception-BN~\cite{ioffe2015batch}.
We train our networks using mini-batch stochastic gradient descent with momentum, as implemented in the Caffe library, using the following meta-parameters: weight decay $5\times 10^{-4}$, momentum $0.9$, initial learning rate $10^{-3}$.
We augment the input data by scaling all images to $256\times 256$ pixels, randomly cropping $227\times 227$ pixels (for AlexNet) or $224\times 224$ pixels (Inception-BN) patches and performing random flips.
In all experiments we choose the parameter $\lambda$, which is fixed for tests of a given setting, by cross-validation.

AlexNet~\cite{krizhevsky2012imagenet} is a well-know architecture with five convolutional and three fully-connected layers, denoted as \texttt{fc6}, \texttt{fc7} and \texttt{fc8}.
The outputs of \texttt{fc6} and \texttt{fc7} are commonly used in the domain-adaptation literature as pre-trained feature representations~\cite{donahue2014decaf,sun2016return} for traditional machine learning approaches.
In our experiments we modify AlexNet by appending a \DAL to each fully-connected layer.
Differently from the original AlexNet, we \emph{do not} perform dropout on the outputs of \texttt{fc6} and \texttt{fc7}.
We initialize the network parameters from a publicly-available model trained on the ILSVRC-2012 data, we finetune all layers, and learn from scratch the last \texttt{fc} layer (we increase its learning rate by a factor of $10$).
During training, each mini-batch contains a number of source and target samples proportional to the size of the corresponding dataset, while the batch size remains fixed at 256.
We train for a total of 60 epochs (where ``epoch'' refers to a complete pass over the source set), reducing the learning rate by a factor $10$ after 54 epochs.

Inception-BN~\cite{ioffe2015batch} is a very deep architecture obtained by concatenating ``inception'' blocks.
Each block is composed of several parallel convolutions with batch normalization and pooling layers.
To apply the proposed method to Inception-BN, we replace each batch-normalization layer with a \DAL.
Similarly to AlexNet, we initialize the network's parameters from a publicly-available model trained on the ILSVRC-2012 data and freeze the first three inception blocks.
Each batch is composed of 32 source images and 16 target images.
In the Office-31 experiments we train for 20 epochs, reducing the learning rate by a factor 10 every $33\,\%$ of the total number of iterations.
\vspace{-0.4cm}
\subsubsection{\DIAL variations.}
\label{sec:variations}  
To evaluate the robustness of our framework, we tested the $3$ \DIAL variations we discussed in Section \ref{sec:predictors}:
 classical Batch Normalization, reported as \textit{BN}, 
 Batch Normalization with prior on variance, reported as \textit{Epsilon}\footnote{The $\epsilon$ parameter is set to $1$ for all experiments.},
 Laplacian Batch Normalization, reported as \textit{Laplacian BN}.
 
Furthermore, we also tested a new sparse regularizer that has been recently proposed in~\cite{ren2016normalizing}, which operates at level of the centered features in the batch-normalization layer (before normalization by the variance). This is beneficial in terms of decorrelating the features and can be integrated readily in our framework. We consider the new regularizer for our \DALs that are based on batch-normalization and regard them as Batch Normalization with sparsity, reported as \textit{sparse} and Batch Normalization with prior on variance and sparsity, reported as \textit{Epsilon sparse}.

%Each of these $5$ settings was tested with and without the use of an Entropy loss.
%, while in the Cross-Dataset Testbed experiments we train for 25 epochs, reducing the learning rate every 10 epochs.

%% file: results.tex
\vspace{-0.4cm}
\subsection{Results}
\vspace{-0.2cm}
\subsubsection{Comparison with State-of-the art Methods.}
\label{sec:comparison}

\begin{table*}[t]
  \centering
  \begin{tabular}{lrccccccc}  
  \toprule
  \multirow{2}{*}{Method} & \small{Source} & Amazon & Amazon & Webcam & Webcam & DSLR   & DSLR   & Average\\
                          & \small{Target} & Webcam & DSLR   & Amazon & DSLR   & Amazon & Webcam &\\
  \midrule
%  \multicolumn{2}{l}{TCA~\cite{pan2011domain}} 
%    & $21.5$ & $11.4$ & $14.6$ & $58.4$ & $8.0$ & $50.1$  & $27.3$\\
%  \multicolumn{2}{l}{GFK~\cite{gong2012geodesic}} 
%    & $19.7$ & $10.6$ & $15.8$ & $63.1$ & $7.9$ & $49.7$  & $27.8$\\  
%   \multicolumn{2}{l}{GFK -- Inception-BN~\cite{gong2012geodesic}} 
%     & $66.70$ & $70.10$ & $56.90$ & $99.40$ & $58.00$ & $97.00$  & $74.7$\\
%   \multicolumn{2}{l}{SA -- Inception-BN~\cite{fernando2013unsupervised}} 
%     & $69.80$ & $71.30$ & $56.90$ & $99.00$ & $59.40$ & $95.50$ & $75.3$\\
%  \midrule
  \multicolumn{2}{l}{AlexNet -- source~\cite{krizhevsky2012imagenet}} 
    & $61.6$ & $63.8$ & $49.8$ & $99.0$ & $51.1$ & $95.4$ & $70.1$ \\
  \multicolumn{2}{l}{DDC~\cite{tzeng2014deep}} 
    & $61.8$ & $64.4$ & $52.2$ & $98.5$ & $52.1$ & $95.0$ & $70.6$\\
  \multicolumn{2}{l}{DAN~\cite{long2015learning}} 
    & $68.5$ & $67.0$ & $53.1$ & $99.0$ & $54.0$ & $96.0$ & $72.9$\\
  \multicolumn{2}{l}{ReverseGrad~\cite{ganin2014unsupervised}} 
    & $73.0$ & -- & -- & $99.2$ & -- & $96.4$ & -- \\
  %\multicolumn{2}{l}{\DIALAlexNet}
  %  & $73.2$ & $71.7$ & $56.2$ & $99.3$ & $59.6$ & $95.9$ & $76.0$\\
  \multicolumn{2}{l}{\DIALAlexNet $sparse$}
    & $76.5$ & $72.4$ & $55.9$ & $99.4$ & $58.6$ & $97.0$ & $76.5$\\
  %\multicolumn{2}{l}{\DIALAlexNet-No DA}
   % & $62.7$ & $66.8$ & $45.4$ & $99.4$ & $42.3$ & $95.4$ & $68.7$\\
  %\multicolumn{2}{l}{\DIALAlexNet}
   % & $73.2$ & $71.7$ & $57.2$ & $99.3$ & $59.6$ & $95.9$ & $76.2$\\
  \midrule
  \multicolumn{2}{l}{Inception-BN -- source~\cite{ioffe2015batch}} 
    & $70.3$ & $70.5$ & $57.9$ & \boldmath{$100.0$} & $60.1$ & $94.3$ & $75.5$\\
  \multicolumn{2}{l}{AdaBN~\cite{li2016revisiting}}
    & $74.2$ & $73.1$ & $57.4$ & $99.8$ & $59.8$ & $95.7$ & $76.7$\\
  \multicolumn{2}{l}{AdaBN + CORAL~\cite{li2016revisiting}}
    & $75.4$ & $72.7$ & $60.5$ & $99.6$ & $59.0$ & $96.2$ & $77.2$\\
  \multicolumn{2}{l}{\DIALInception $BN$}
    & \boldmath{$82.9$} & \boldmath{$87.3$} & \boldmath{$62.6$} & $99.9$ & \boldmath{$63.1$} & \boldmath{$98.2$} & \boldmath{$82.4$}\\  
  \bottomrule
  \end{tabular}
  \vspace{0.4em}
  \caption{Results on the Office-31 dataset using the full protocol.}
  \label{tab:office-31}
    \vspace{-0.9cm}
\end{table*}
%\vspace{-0.2cm}
In our first series of experiments, summarized in Table~\ref{tab:office-31}, we compare our approach, applied to both AlexNet and Inception-BN, with several state-of-the-art methods on the Office-31 dataset.
In particular, we consider: several deep methods based on AlexNet-like architectures, \ie  Deep Adaptation Networks (DAN) \cite{long2015learning}, Deep Domain Confusion (DDC) \cite{tzeng2014deep}, the ReverseGrad network \cite{ganin2014unsupervised}; a recent deep method based on the Inception-BN architecture, \ie AdaBN~\cite{li2016revisiting} with and without CORAL feature alignment \cite{sun2016return}.
We compare these baselines to the AlexNet and Inception-BN networks modified with our approach as explained in Section~\ref{sec:setup}, reporting the best results among the \DAL variations we experimented with (see Table.~\ref{tab:office-31-ablation}).
In the table our approach is denoted as \DIALAlexNet and \DIALInception.
As a reference, we further report the results obtained considering standard AlexNet and Inception-BN networks trained only on source data.

Among the deep methods based on the AlexNet architecture, \DIALAlexNet shows the best average performance.
%On the other hand, the results obtained with \DIALAlexNet-No DA are the lowest among the AlexNet-derived architectures, highlighting the effectiveness of the proposed domain-alignment technique.
%coupling both components of the proposed method.
Among the methods based on Inception-BN, our approach considerably outperforms the others, with an average accuracy of five points higher than the second best, and improvements on the single experiments as high as ten points.
It is interesting to note that the relative increase in accuracy from the source-only Inception-BN to \DIALInception is higher than that from the source only AlexNet to \DIALAlexNet.
The considerable success of our method in conjunction with Inception-BN can be attributed to the fact that, differently from AlexNet, this network is pre-trained with batch normalization, and thus initialized with weights that are already calibrated for normalized features.
%In appendix ~\ref{sec:office31}, we also provide \DIALAlexNet results on Office 31, using the classical\cite{saenko2010adapting} sampling protocol and show that our method can successfully deal with less samples.
\vspace{-0.4cm}
\subsubsection{In-depth analysis of \DALs.}
\label{sec:ablation}

\begin{table*}[t]
  \centering
  \begin{tabular}{lrccccccc}  
  \toprule
  \multirow{2}{*}{Method} & \small{Source} & Amazon & Amazon & Webcam & Webcam & DSLR   & DSLR   & Average\\
                          & \small{Target} & Webcam & DSLR   & Amazon & DSLR   & Amazon & Webcam &\\
    \midrule
    \multicolumn{9}{c}{\emph{Baselines}} \\
    \multicolumn{2}{l}{AlexNet -- source~\cite{krizhevsky2012imagenet}} 
    & $61.6$ & $63.8$ & $49.8$ & $99.0$ & $51.1$ & $95.4$ & $70.1$ \\
    \multicolumn{2}{l}{AlexNet -- Entropy loss} 
    & $63.7$ & $65.6$ & $35.5$ & $96.6$ & $42.9$ & $99.6$ & $67.3$ \\
    \midrule
    \multicolumn{9}{c}{\emph{With entropy loss}} \\
    \multicolumn{2}{l}{\DIALAlexNet $BN$}
    & $73.2$ & $71.7$ & $56.2$ & $99.3$ & $59.6$ & $95.9$ & $76.0$\\
    \multicolumn{2}{l}{\DIALAlexNet $Epsilon$}
    & $71.6$ & $71.7$ & $56.7$ & $99.3$ & $59.4$ & $99.2$ & $76.3$\\
    \multicolumn{2}{l}{\DIALAlexNet $sparse$}
    & $76.5$ & $72.4$ & $55.9$ & $99.4$ & $58.6$ & $97.0$ & $76.5$\\
    \multicolumn{2}{l}{\DIALAlexNet $Epsilon$ $sparse$}
    & $72.1$ & $72.3$ & $57.0$ & $99.7$ & $59.0$ & $97.2$ & $76.2$\\
    \multicolumn{2}{l}{\DIALAlexNet $Laplacian$ $BN$}
    & $73.0$ & $72.0$ & $55.1$ & $98.7$ & $56.7$ & $96.6$ & $75.4$\\
    \midrule
    \multicolumn{9}{c}{\emph{Without entropy loss}} \\
    \multicolumn{2}{l}{\DIALAlexNet $BN$}
    & $62.2$ & $65.5$ & $47.1$ & $99.2$ & $47.6$ & $95.2$ & $69.5$\\
    \multicolumn{2}{l}{\DIALAlexNet $Epsilon$}
    & $65.3$ & $64.5$ & $47.3$ & $99.5$ & $48.4$ & $95.0$ & $70.0$\\
    \multicolumn{2}{l}{\DIALAlexNet $sparse$}
    & $60.6$ & $64.0$ & $47.0$ & $99.3$ & $48.1$ & $95.6$ & $69.1$\\
    \multicolumn{2}{l}{\DIALAlexNet $Epsilon$ $sparse$}
    & $64.6$ & $65.3$ & $46.9$ & $99.7$ & $48.4$ & $95.7$ & $70.1$\\
    \multicolumn{2}{l}{\DIALAlexNet $Laplacian$ $BN$}
    & $61.8$ & $65.3$ & $46.8$ & $98.4$ & $46.8$ & $94.8$ & $69.0$\\
  \bottomrule
  \end{tabular}
  \vspace{0.2cm}
  \caption{Analysis of the different variants of the proposed DA layers on the Office-31 dataset using the full protocol.}
  \label{tab:office-31-ablation}
    \vspace{-1.2cm}
\end{table*}

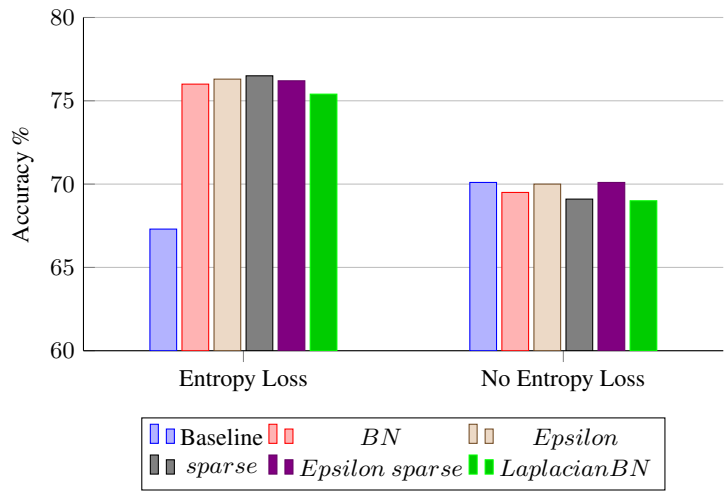
\begin{figure*}
  \centering
  \begin{tikzpicture}
  \begin{axis}[%
      title={}, width=10cm, height=6cm,
      symbolic x coords={da,noda},
      enlarge x limits=0.5, xtick=data,
      xticklabels={Entropy Loss, No Entropy Loss},
      ybar, ymin=60, ymax=80, ylabel={Accuracy \%}, ymajorgrids=true,
      axis lines*=left,
      legend style={at={(0.5,-0.2)},anchor=north,legend columns=3}]
    \addplot coordinates {(da,67.3) (noda,70.1)};
    \addplot coordinates {(da,76.0) (noda,69.5)};
    \addplot coordinates {(da,76.3) (noda,70.0)};
    \addplot coordinates {(da,76.5) (noda,69.1)};
    \addplot coordinates {(da,76.2) (noda,70.1)};
    \addplot coordinates {(da,75.4) (noda,69.0)};
    \legend {Baseline,$BN$,$Epsilon$,$sparse$,$Epsilon$ $sparse$,$Laplacian BN$};
  \end{axis}
  \end{tikzpicture}
  \caption{Comparison of the different variants of the proposed method on the Office-31 dataset (average accuracy across different transfer tasks)}
  \label{fig:ablation}
  \vspace{-0.4cm}
\end{figure*}

In our second series of experiments we aim to characterize the effects of different variations of the proposed \DALs.
To do this, we perform an ablation study considering all possible combinations of the following network variations: (i) with and without the entropy term on the target samples in the loss function; (ii) with and without \DALs; (iii) with the \DAL variations (Sec. \ref{sec:variations}). 

The results are reported in Table~\ref{tab:office-31-ablation}, and further synthesized in Figure~\ref{fig:ablation}.
As anticipated in the previous section, the \textit{\DIALAlexNet sparse} variant achieves the best accuracy.
Overall, independently from the particular \DAL variant, the networks utilizing our proposal in its full extent (\ie those in the ``With entropy loss'' section of Table~\ref{tab:office-31-ablation}) consistently outperform the others, further confirming the validity of our domain adaptation approach.

From the results in Table~\ref{tab:office-31-ablation}, we see that the use of an entropy loss term by itself does not provide any advantage over the baseline approach.
On the contrary, an average drop in accuracy of about $3\,\%$ is observed when comparing AlexNet -- Entropy loss to AlexNet -- source, with partial results greatly varying depending on the particular source / target pair.
Interestingly, AlexNet -- Entropy loss shows better accuracy compared to AlexNet -- source in all the settings in which the target dataset is smaller than the source dataset, \ie A$\rightarrow$W, A$\rightarrow$D and D$\rightarrow$W. This may be explained by the fact that the entropy term is more effective when there are sufficient source samples to appropriately bias the decision boundary.
%\lorenzo{any explanation for this?}
As shown in Figure~\ref{fig:ablation}, the best performance between the proposed variations of our domain alignment layers are obtained when considering BN with sparse activations. % we observe a noticeable performance gap between the two in our experiments. 
Adding a sparse regularizer on the activations helps to decorrelate the filter responses \cite{ren2016normalizing} and our results demonstrate that it has a positive effect on domain adaptation tasks.

%exact computations performed in \DIAL Laplacian, involving sample median and sum of absolute differences, introduce discontinuities in the network, both in the forward and backward passes, which can negatively impact convergence.
%On the other hand, \DIAL L1 loss produces an overall smoother and easier to optimize objective function.

%As mentioned in Section\ref{sec:predictors}, both \DIAL Laplacian and \DIAL L1 loss are derived from the same Laplacian prior on the reference distribution.
%However, as shown in Figure~\ref{fig:ablation}, we observe a noticeable performance gap between the two in our experiments.
%A possible explanation for this lies in the fact that the exact computations performed in \DIAL Laplacian, involving sample median and sum of absolute differences, introduce discontinuities in the network, both in the forward and backward passes, which can negatively impact convergence.
%On the other hand, \DIAL L1 loss produces an overall smoother and easier to optimize objective function.